\documentclass[final,1p,times]{elsarticle}

\usepackage{pifont}
\usepackage{graphicx}
\usepackage{booktabs}
\usepackage{amsmath,amssymb,amsfonts}
\usepackage{algorithm}
\usepackage{algorithmic}
\usepackage{textcomp}
\usepackage{multirow}
\usepackage{hhline, booktabs}
\usepackage{xcolor}
\usepackage{colortbl}
\usepackage{float} 
\usepackage{caption} 
\usepackage{url}
\usepackage{enumerate}
\usepackage{comment}
\usepackage{tikz}
\usepackage{enumitem}
\usepackage{mathabx}
\usepackage{adjustbox}
\setlist[itemize]{left=0pt, itemsep=0pt, topsep=0pt}

\newcommand{\cmark}{\ding{51}}%
\newcommand{\xmark}{\ding{55}}%

\definecolor{promptblue}{RGB}{15, 117, 188}
\definecolor{promptgreen}{RGB}{7,117,131}
\definecolor{promptpurple}{RGB}{166, 128, 184}

\definecolor{mycolor}{RGB}{187, 191, 188}
\definecolor{softcolortable}{RGB}{242, 242, 242}
\definecolor{ForestGreen}{RGB}{34,139,34}

\definecolor{blue}{RGB}{151, 208, 119}
\definecolor{orange}{RGB}{181, 115, 157}
\definecolor{green}{RGB}{126, 166, 224}
\definecolor{red}{RGB}{234, 107, 102}
\definecolor{purple}{RGB}{255, 181, 112}
\definecolor{brown}{RGB}{255, 217, 102}
\definecolor{pink}{RGB}{241, 156, 153}

\newcommand{\improvement}[1]{\textcolor{ForestGreen}{(+#1)}}

\usepackage{tikz}
\usepackage{lipsum}
\usepackage{makecell}

\newlength{\RoundedBoxWidth}
\newsavebox{\GrayRoundedBox}
\newenvironment{GrayBox}[1][\dimexpr0.48\textwidth-4.5ex]%
   {\setlength{\RoundedBoxWidth}{\dimexpr#1}
    \begin{lrbox}{\GrayRoundedBox}
       \begin{minipage}{\RoundedBoxWidth}}%
   {   \end{minipage}
    \end{lrbox}
    \begin{center}
    \begin{tikzpicture}%
       \draw node[draw=black,fill=black!10,%
             inner sep=3mm, outer sep=0mm, text width=\RoundedBoxWidth, align=left, rounded corners=3mm]%
             {\usebox{\GrayRoundedBox}};
    \end{tikzpicture}
    \end{center}}

\journal{CSBJ}

\begin{document}

\begin{frontmatter}




\title{A Two-Step Concept-Based Approach for Enhanced Interpretability and Trust in Skin Lesion Diagnosis}


\author[label1,label3]{Cristiano Patrício} 
\author[label2,label3]{Luís F. Teixeira} 
\author[label1]{João C. Neves} 

\affiliation[label1]{organization={Universidade da Beira Interior and NOVA LINCS},
            country={Portugal}}

\affiliation[label2]{organization={Faculdade de Engenharia da Universidade do Porto},
country={Portugal}}

\affiliation[label3]{organization={INESC TEC},
country={Portugal}}

\begin{abstract}
The main challenges hindering the adoption of deep learning-based systems in clinical settings are the scarcity of annotated data and the lack of interpretability and trust in these systems. Concept Bottleneck Models (CBMs) offer inherent interpretability by constraining the final disease prediction on a set of human-understandable concepts. However, this inherent interpretability comes at the cost of greater annotation burden. Additionally, adding new concepts requires retraining the entire system. In this work, we introduce a novel two-step methodology that addresses both of these challenges. By simulating the two stages of a CBM, we utilize a pretrained Vision Language Model (VLM) to automatically predict clinical concepts, and an off-the-shelf Large Language Model (LLM) to generate disease diagnoses based on the predicted concepts. Furthermore, our approach supports test-time human intervention, enabling corrections to predicted concepts, which improves final diagnoses and enhances transparency in decision-making. We validate our approach on three skin lesion datasets, demonstrating that it outperforms traditional CBMs and state-of-the-art explainable methods, all without requiring any training and utilizing only a few annotated examples. The code is available at \url{https://github.com/CristianoPatricio/2-step-concept-based-skin-diagnosis}.
\end{abstract}




\end{frontmatter}



\section{Introduction}
\label{sec:intro}


Recent advancements in automated computer-aided diagnosis systems for detecting diseases from medical images have demonstrated a significant improvement in performance, largely driven by the superior capabilities of deep learning models. This shift has considerably boosted the accuracy of these systems in delivering precise diagnoses across various medical imaging tasks, including skin lesion analysis, sometimes achieving results comparable to those of dermatologists~\cite{codella2017deep, esteva2017dermatologist}. Despite these advancements, state-of-the-art deep learning-based systems in dermatology operate as ``black-boxes'', failing to provide explainable rationale behind the decision-making process. This lack of transparency remains a major obstacle to their widespread adoption and integration into clinical practice~\cite{rotemberg2019role}.

To address this issue, explainable models have been developed to enhance the transparency and reliability of medical AI systems. Among these, Concept Bottleneck Models (CBMs)\cite{koh_2020_ICML} have gained considerable attention in medical imaging analysis\cite{Fang_MM2020,Lucieri_IJCNN2020,Patricio_2023_CVPRW}, as they base their decisions on the presence or absence of human-interpretable concepts, closely resembling the way clinicians analyze medical images. Studies have also shown that concept-based explanations are generally preferred by humans over other methods, such as heatmaps or example-based explanations~\cite{ramaswamy2023overlooked}. However, despite their growing popularity, CBMs face two key challenges: i) they require detailed annotations of human-understandable concepts, a process that is both time-consuming and demands domain expertise; and ii) the model must be retrained whenever new concepts are introduced. While recent studies~\cite{yang2023language,oikarinen2023label,menon2022visual,yan2023robust} have attempted to reduce the reliance on concept annotations by leveraging Large Language Models (LLMs) to generate candidate concepts, these approaches still require retraining when new target classes or concepts are added to the system.

\begin{figure*}[t]
    \centering
    \includegraphics[width=\textwidth]{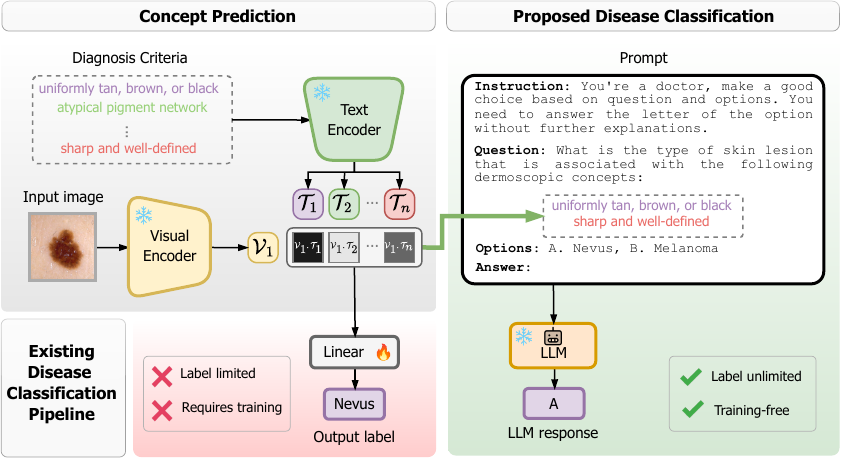}
    \caption{\textbf{Overview of the proposed framework}. The linear classifier layer (left) is replaced by a pretrained Large Language Model (LLM) (right), which grounds its responses on clinical concepts predicted by a pretrained vision-language model (VLM). This approach is training-free and not restricted by predefined labels, allowing the LLM to generate diverse diagnostic possibilities for different diseases.}
    \label{fig:method}
\end{figure*}

In this work, we address the limitations of CBMs by proposing a novel two-step approach that provides concept-based explanations and generates disease diagnoses grounded in predicted concepts, all without the need for additional training. In the first stage, we utilize pretrained Vision-Language Models (VLMs) to predict the presence of a set of clinical concepts. In the second stage, these predicted concepts are incorporated into a custom designed prompt, which is used to query an LLM for a final diagnosis class to ensure the diagnosis is based on the identified concepts. Figure \ref{fig:method} illustrates our methodology compared to the traditional approach where a linear layer is trained to predict the final target class. Similar to CBMs, our approach allows for test-time intervention, enabling clinicians to correct concept predictions at inference time. However, since the predicted concepts by VLM are presented as plain text for the final diagnosis, this enables more intuitive human intervention compared to adjusting numerical values in the bottleneck layer, as in traditional CBM test-time interventions. Additionally, leveraging a VLM to predict concepts offers the advantage of incorporating both visual and textual information during model inference. In the other hand, unlike CBMs, our approach does not require training to provide the final diagnosis class and can be easily adapted to incorporate new concepts. We validate our approach on three well-known skin lesion datasets, demonstrating that it outperforms traditional CBMs and state-of-the-art explainable methods, all without requiring any training and using only a few annotated examples within the prompt. Our contributions are as follows: 

\begin{itemize}
    \item We introduce a novel two-step approach for concept-based explainability in skin lesion diagnosis without requiring additional training;
    \item We reduce the annotation burden of CBMs by leveraging zero-shot capabilities of pretrained VLMs to automatically predict clinical concepts;
    \item We provide the final diagnosis class using tailored prompts and leveraging the few-shot capabilities of LLMs;
    \item Our approach enables human test-time intervention, allowing corrections to any miscpredicted concepts;
    \item We demonstrate superior interpretability and performance compared to traditional CBMs on three public benckmarks.
\end{itemize}

The remainder of the paper is organized as follows: Section \ref{sec:related_work} reviews related work in the context of our study. Section \ref{sec:methodology} introduces the proposed two-step methodology. Section \ref{sec:experimental_results} outlines the experimental setup and implementation details. Section \ref{sec:results} presents key results and analysis. A general discussion is provided in Section \ref{sec:general_discussion}. Finally, Section \ref{sec:limitations} and Section \ref{sec:conclusion} discuss the limitations and conclude the paper, respectively.

\section{Related Work}
\label{sec:related_work}

\subsection{Vision-Language Models}

Vision-language models (VLMs) aim to learn joint representations from image-text pairs using contrastive learning. While models like CLIP~\cite{radford2021learning} have demonstrated impressive zero-shot capabilities across a wide range of tasks, their application in the biomedical domain remains challenging due to distribution shifts and the domain specific vocabulary. To address these challenges, several medical-specific VLMs have been developed, such as BiomedCLIP~\cite{zhang2023biomedclip}, PubMedCLIP~\cite{eslami2023pubmedclip}, and MedCLIP~\cite{wang2022medclip}, which are pretrained on large-scale image-text pairs from sources like PubMed articles and radiology datasets.
However, despite these advancements, such models still underperform when compared to task-specific models~\cite{patricio2024towards} and lack transparency and interpretability in their decision-making processes. Our novel methodology leverages the zero-shot capabilities of medical-domain VLMs to automatically annotate concepts, forming the first step of our approach.


\subsection{Explainable Models for Skin Lesion Diagnosis}

Concept Bottleneck Models (CBMs)~\cite{koh_2020_ICML} first predict a set of human-interpretable concepts and then use these concepts to determine the final class label using a linear layer, thus offering transparency in the decision-making process. However, these models require extensive manual concept annotations and depend on a linear layer for the final diagnosis prediction. To address the need for annotations, several studies~\cite{yang2023language,oikarinen2023label,menon2022visual,yan2023robust} have explored the use of Large Language Models (LLMs) to generate candidate concepts for target classes. Other methods~\cite{patricio2024towards, kim2024transparent, gao2024aligning} align visual features with pre-defined clinically relevant concepts using vision-language models (VLMs). 

Despite these improvements, most models still rely on a linear classifier to predict the final diagnostic label, whether based on concepts or visual features. Our approach overcomes this by prompting an LLM to directly predict the diagnosis using a tailored prompt that incorporates the concepts extracted by a pretrained VLM. This eliminates the fixed label constraint, improving scalability and removing the need for retraining when new diagnostic categories or concepts are introduced.




\section{Methodology}
\label{sec:methodology}

Given the task of predicting a target disease $y \in \mathbb{R}$ from input $x \in \mathbb{R}^d$, let $\mathcal{D} = \{(x^{(i)}, y^{(i)}, c^{(i)})\}_{i=1}^{n}$ represent a batch of training samples, where $c \in \mathbb{R}^l$ is a vector of $l$ clinical concepts. CBMs first map the input $x$ into a set of interpretable concepts $c$ (the ``bottleneck'') by learning a function $g: \mathbb{R}^d \rightarrow \mathbb{R}^l $ ($x \rightarrow c$), and use these concepts to predict the target $y$ through $f: \mathbb{R}^l \rightarrow \mathbb{R}$ ($c \rightarrow y$). As a result, the final prediction $\hat{y} = f(g(x))$ is entirely based on the predicted concepts $\hat{c} = g(x)$.
To estimate the presence of clinical concepts from an input image, we utilize a pretrained VLM and calculate the cosine similarity between the image features and a set of clinical concepts in the concept set $C = \{c^{(0)},...,c^{(i)}\}$ (see Section \ref{subsec:concepts_prediction}). Next, we prompt a pretrained LLM to generate the final diagnosis class grounded on the predicted concepts (detailed in Section \ref{subsec:disease_classification}). Figure~\ref{fig:method} provides an overview of our methodology, which is further described in the following subsections.

\subsection{Concept Prediction}
\label{subsec:concepts_prediction}

The first stage focuses on predicting the presence of dermoscopic concepts from a given input image, representing the ``bottleneck'' of CBM: $x \rightarrow c$. Following previous work~\cite{patricio2024towards}, we adopt a pretrained VLM and determine the presence of a dermoscopic concept $c$ in the input image by assessing the similarity between the image feature embedding, $\mathcal{V}(x)$, and the feature embedding of each concept $c$. Formally, the similarity scores are given by:
\begin{equation}
\label{eq:similarity_scores}
    scores = sim(\mathcal{V}(x), \mathcal{T}(c^{(i)})),
\end{equation}
where $sim(.)$ is a similarity metric (e.g., cosine similarity), $\mathcal{V}(.)$ is the visual encoder, and $\mathcal{T}(.)$ is the text encoder.

\subsection{Disease Classification}
\label{subsec:disease_classification}

In the second stage, we prompt a Large Language Model (LLM) to generate the final disease diagnosis grounded on the predicted dermoscopic concepts, corresponding to the classification phase of CBM: $c \rightarrow y$. Using the concept scores generated in the first stage (Equation \ref{eq:similarity_scores}), we binarize them using a threshold and map them to their respective concept names. These concepts are then incorporated into the designed prompt. An example of this prompt is provided on the right side of Figure \ref{fig:method}. This method ensures that the diagnosis is based on the dermoscopic concepts, improving the interpretability and transparency of the model's output, and eliminates the need for training a linear layer, allowing for more flexible and varied diagnostic output formats.

\subsection{Few-Shot Prompting}
\label{subsec:few_shot_prompting}

Few-shot prompting, also known as In-Context Learning (ICL), is a prompt engineering technique where the model is given with a few task demonstrations during inference within the prompt as conditioning~\cite{radford2019language}. This method has been shown to significantly boost the performance of LLMs~\cite{gpt3}. The process involves presenting the model with $K$ demonstration examples, containing the context and ground truth answers, followed by a final example (query), where the model is expected to generate the answer following the template of the given demonstrations. The key advantage of this approach is that it requires no updates to the model's weights. An example of such a prompt is shown in Figure \ref{fig:prompt_demonstration_examples}.
\begin{figure*}
    \centering
    \begin{GrayBox}[0.9\textwidth]
    {{\textit{\textbf{Instruction}}}} \\
    You're a doctor, make a good choice based on the question and options. You need to answer the letter of the option without further explanations. \\ \\
    {{\textit{\textbf{Demonstrations}}}} \\
    Consider the following examples: \\ 
    1. The color is highly variable, often with multiple colors (black, brown, red, white, blue), the shape is irregular, the border is often blurry and irregular, the dermoscopic patterns are atypical pigment network, irregular streaks, blue-whitish veil, irregular, the texture is a raised or ulcerated surface, the symmetry is asymmetrical, the elevation is flat to raised. Thus the diagnosis is \textcolor{promptblue}{melanoma}. \\
    2. The color is uniformly tan, brown, or black, the shape is round, the border is sharp and well-defined, the dermoscopic patterns are regular pigment network, symmetric dots and globules, the texture is smooth, the symmetry is symmetrical, the elevation is raised with possible central ulceration. Thus the diagnosis is \textcolor{promptblue}{nevus}.
    \\
    (...)
    \\
    \\
    {{\textit{\textbf{Question}}}} \\
    What is the type of skin lesion that is associated with the following dermoscopic concepts ?\\
    The color is uniformly tan, brown, or black, the shape is round, the border is sharp and well-defined, the dermoscopic patterns are regular pigment network, symmetric dots and globules, the texture is smooth, the symmetry is symmetrical, the elevation is raised with possible central ulceration.
    \\
    \\
    {{\textit{\textbf{Options}}}} \\
    A. Nevus \\
    B. Melanoma \\
    \textcolor{promptblue}{Answer:}
    \end{GrayBox}
    \caption{\textbf{Example prompt with $K=2$ demonstration examples.} Few-shot prompting is expected to improve performance as the number of demonstration examples increases.}
    \label{fig:prompt_demonstration_examples}
\end{figure*}
While the demonstrations could be selected randomly, Zhou et al.~\cite{zhou2024adapting} have shown that this approach can lead to decreased model performance compared to the zero-shot scenario. To address this, following the method in~\cite{yang2022empirical}, we adopt a selection strategy where demonstrations are chosen based on the similarity of the query image $x$ to a bank of precomputed image features from training set. Specifically, given the feature vector $\mathcal{V}(x)$ of the query image, we select the $N$ most similar image feature vectors from the training set $\mathcal{D}$. Let $S$ denote the set of indices corresponding to the $N$ most similar feature vectors:
\begin{equation}
     S = \text{argsort}\left(\{ sim(\mathcal{V}(x), f_i) \}_{i=0}^n \right)[:N],
\end{equation}
where $sim(.)$ represents a similarity metric (e.g., cosine similarity), $f_i$ the image feature from the training set, and $\mathcal{V}(.)$ is an arbitrary visual encoder. Once the indices $S$ of the similar images are identified, we retrieve their corresponding ground truth concepts and class labels from the training set. These are used to create demonstration examples, which are then appended to the prompt, as illustrated in Figure \ref{fig:prompt_demonstration_examples}.

\section{Experiments}
\label{sec:experimental_results}

In this section, we outline our experimental setup, including details of the adopted models, datasets, and evaluation metrics, as well as the implementation details.

\subsection{Setup}

\vspace{2mm}
\paragraph{\textbf{Models}}
In our experiments, we evaluate the performance of our approach against the standard CBM~\cite{koh_2020_ICML} and several pretrained Vision-Language Models (VLMs), including the generic CLIP~\cite{radford2021learning}, the medical-domain BiomedCLIP~\cite{zhang2023biomedclip}, and three dermatology-specific VLMs: MONET~\cite{kim2024transparent}, CBI-VLM~\cite{patricio2024towards}, and ExpLICD~\cite{gao2024aligning}. For the Large Language Models (LLMs), we compare the generic Mistral-7B-Instruct~\cite{jiang2023mistral}, the medical-domain MMed-LLama3-8B-EnIns~\cite{qiu2024towards}, and conduct additional experiments with GPT-4~\cite{openai2023gpt4} in the top-performing setting.

\paragraph{\textbf{Datasets}}
We conduct experiments using three publicly available dermoscopic datasets: \textbf{PH$^2$}~\cite{PH2}, \textbf{Derm7pt}~\cite{DERM7PT}, and \textbf{HAM10000}~\cite{tschandl2018ham10000}. The choice of these datasets was determined by the availability of clinical concept annotations. The PH$^2$ dataset consists of dermoscopic images of melanocytic lesions, including ``melanoma'' and two types of ``nevus'', which we merged and treated as a single ``nevus'' category. For this dataset, we applied 5-fold cross-validation due to its small size. Derm7pt contains  over 2,000 clinical and dermoscopic images, which we filtered to include only images of ``nevus'' and ``melanoma'' classes. HAM10000 contains 10,015 dermoscopic images of various skin lesions across different body locations. Following~\cite{chanda2024dermatologist}, we used all biopsy-verified ``melanoma'' and ``nevus'' images from HAM10000, resulting in a set of $n = 3,445$ images. Detailed statistics, including the train/validation/test splits for each dataset, are shown in Table \ref{tab:dataset_statistics}. All reported results are based on the entire test set for each dataset. For Derm7pt, we followed the official split partition, and for HAM10000, we used the partitioning described in~\cite{chanda2024dermatologist}.

\begin{table}[!ht]
  \centering
  \setlength{\tabcolsep}{5pt}
  \begin{adjustbox}{width=0.9\textwidth}
  \begin{tabular}{lccccc}
    \toprule
     \textbf{Dataset} & \textbf{Classes} & \textbf{Train size} & \textbf{Validation size} & \textbf{Test size} & \textbf{\% (MEL/NEV)} \\
    \midrule
    PH$^2$~\cite{PH2} & 2 & 160 (28 to 34) & - & 40 (6 to 12) & 25/75\\
    Derm7pt~\cite{DERM7PT} & 2 & 346 (90) & 161 (61) & 320 (101) & 30/70\\
    HAM10000~\cite{tschandl2018ham10000} & 2 & 2,646 (752) & 599 (172) & 200 (100) & 30/70\\
    \bottomrule
  \end{tabular}%
\end{adjustbox}
  \caption{\textbf{Dataset statistics}. Numbers between rounded brackets represent the $\#$ of Melanoma examples in the split. MEL is Melanoma and NEV is Nevus.}
  \label{tab:dataset_statistics}
\vspace{-0.6cm}
\end{table}

\paragraph{\textbf{Evaluation Metrics}}
\label{subsec:metrics}

We adopt Balanced Accuracy (BAcc) as the the arithmetic mean of Sensitivity (TPR) and Specificity (TNR), as this metric is tailored to deal with imbalanced datasets, which is the case of the considered datasets (see \textbf{\%(MEL/NEV)} column of Table \ref{tab:dataset_statistics}).

\subsection{Implementation Details}
\label{subsec:implementation_details}

For experiments with concept-based models, we reproduce the results using the official implementation of CBM~\cite{koh_2020_ICML}.
For pretrained VLMs, we use the checkpoints from the HuggingFace library for CLIP, BiomedCLIP and MONET. Additionally, we employ the official implementations of CBI-VLM and ExpLICD, as provided in~\cite{patricio2024towards} and~\cite{gao2024aligning}, respectively. Checkpoints for Mistral-7B-Instruct and MMed-LLama3-8B-EnIns are also sourced from the HuggingFace library.

For the concept prediction task, we consider the concept ontology of MONET as target set of concepts for experiments with CLIP, BiomedCLIP, MONET and CBI-VLM. For ExpLICD, we utilize its own concept ontology, as the model was fine-tuned with a pre-defined set of concepts, either gathered by LLMs or medical experts. A summary of the ontologies used for each model is provided in Table \ref{tab:concept_ontology}. It is important to note that we only rely on annotated ground-truth concepts for each dataset when performing test-time interventions.

\begin{table*}[!h]
    \centering
    \begin{adjustbox}{width=\textwidth}
        \begin{tabular}{lc}
        \toprule
        
        \multirow{12}{*}{\rotatebox[origin=c]{90}{\large{\textbf{Model-specific}}}}
        & \cellcolor{mycolor} \textbf{\textit{ExpLICD}} \\
        & \textbf{Color}: highly variable, often with multiple colors (black, brown, red, white, blue), \\
        & uniformly tan, brown, or black\\
        &\textbf{Shape}: irregular, round \\
        &\textbf{Border}: often blurry and irregular; sharp and well-defined \\
        &\textbf{Dermoscopic Patterns}: atypical pigment network, irregular streaks, blue-whitish veil, \\ &irregular, regular pigment network, symmetric dots and globules \\
        &\textbf{Texture}: a raised or ulcerated surface; smooth \\
        &\textbf{Symmetry}: asymmetrical; symmetrical \\
        &\textbf{Elevation}: flat to raised \\
        & {\cellcolor{mycolor} \textbf{\textit{MONET}}} \\
        & Asymmetry, Irregular, Black, Blue, White, Brown, Erosion, Multiple Colors, Tiny, Regular \\
        \midrule
        \multirow{8}{*}{\rotatebox[origin=c]{90}{\large{\textbf{Dataset-specific}}}} & {\cellcolor{mycolor} \textbf{\textit{HAM10000}}} \\
        & thick reticular or branched lines; black dots or globules in the periphery of the lesion; \\
        & white lines or white structureless area; eccentrically located structureless area; grey patterns; \\ 
        & polymorphous vessels; pseudopods or radial lines at the lesion margin that do not occupy the entire lesional circumference; \\
        & asymmetric combination of multiple patterns or colours in the absence of other melanoma criteria; melanoma simulator \\
        & \cellcolor{mycolor} \textbf{\textit{PH$^2$ and Derm7pt}} \\
        & typical pigment network, atypical pigment network, irregular streaks, regular streaks, \\
        & regular dots and globules, irregular dots and globules, blue-whitish veil, regression structures \\
        \bottomrule
        \end{tabular}
     \end{adjustbox}
    \caption{\textbf{Concept ontology used in each model and dataset.} Model-specific concepts were collected by either medical experts or LLM and therefore automatically annotated. Dataset-specific concepts are the ground-truth annotations.}
    \label{tab:concept_ontology}
\end{table*}

For the selection strategy used to perform ICL, the image similarity is determined using the cosine similarity between image features, which are extracted from the visual encoder of ExpLICD. Further details, documentation and a demo of our approach are available at \url{https://github.com/CristianoPatricio/2-step-concept-based-skin-diagnosis}. All experiments are run on a single A40 GPU with 48GB of VRAM.

\section{Results and Analysis}
\label{sec:results}

This section presents the key results of our two-step approach for skin lesion classification across three dermoscopic datasets. Specifically, we compare the performance of our approach (\textbf{\textit{VLM + LLM}}) against two baseline approaches:

\begin{itemize}
    \item \textbf{\textit{CBM}}: We use the original CBM, which operates in two stages: i) first, a CNN is used to predict concepts from the input image ($x \rightarrow c$), and ii) a linear model then uses these predicted concepts to infer the final target ($c \rightarrow y$). This approach requires concept annotations and training in both stages.
    \item \textbf{\textit{VLM + Linear Classifier}}: We use a pretrained VLM to predict the concepts from input images (the ``bottleneck'' part), followed by training a linear classifier on the predicted concepts to determine the class label. Training is only needed for the $c \rightarrow y$ phase.
\end{itemize}

Our two-step approach, \textbf{\textit{VLM + LLM}}, leverages a pretrained VLM to predict concepts from an input image (Step 1) and prompts an off-the-shelf LLM to generate the final diagnosis (Step 2) based on the predicted concepts.

\subsection{Disease Classification Performance}

\begin{table}[!t]
  \centering
  \setlength{\tabcolsep}{3.5pt}
  \begin{adjustbox}{width=\textwidth}

  \begin{tabular}{llcccc|ccc|ccc}
    \toprule
     &\multirow{3}{*}{\textbf{Model}} & \multirow{3}{*}{\textbf{Trainable}} &  \multicolumn{9}{c}{\textbf{Dataset}} \\
     \cmidrule{4-12}
     & & & \multicolumn{3}{c}{PH$^2$} & \multicolumn{3}{c}{Derm7pt} & \multicolumn{3}{c}{HAM10000} \\
     & & & \small{Sens.} & \small{Spec.} & \small{BAcc} & \small{Sens.} & \small{Spec.} & \small{BAcc} & \small{Sens.} & \small{Spec.} & \small{BAcc} \\
    \cline{1-12}
    \noalign{\vspace{0.1ex}}
    \rowcolor{mycolor} 
    & \multicolumn{11}{c}{\textbf{\textit{Image to Concepts to Label Classification} $[x \rightarrow c \rightarrow y]$}} \\
    \cline{1-12}
    \noalign{\vspace{0.5ex}}
    & CBM~\cite{koh_2020_ICML} & \cmark & 42.04 & 98.06 & 70.14 & 69.31 & 89.50 & {77.56} & 51.00 & 94.00 & {70.83} \\
    \midrule
    \multirow{5}{*}{\rotatebox[origin=c]{90}{\scriptsize{\textbf{\shortstack{VLM+\\Linear Classifier}}}}} & 
    CLIP~\cite{radford2021learning} + Linear Classifier & \cmark & 59.29 & 96.32 & 77.80 & 37.62 & 91.32 & 64.47 & 21.00 & 94.00 & 57.50 \\
    &BiomedCLIP~\cite{zhang2023biomedclip} + Linear Classifier & \cmark & 22.78 & 98.18 & 60.49 & 6.93 & 99.09 & 53.01 & 0.00 & 100.0 & 50.00 \\
    &MONET~\cite{kim2024transparent} + Linear Classifier & \cmark & 71.43 & 96.79 & \underline{84.11} & 53.47 & 92.69 & {73.08} & 20.00 & 92.00 & 56.00 \\
    &CBI-VLM~\cite{patricio2024towards} + Linear Classifier & \cmark & 24.60 & 98.22 & 61.41 & 50.50 & 89.95 & 70.22 & 56.00 & 92.00 & 74.00 \\
    &ExpLICD~\cite{gao2024aligning} + Linear Classifier & \cmark & 58.49 & 96.47 & 77.48 & 55.45 & 95.43 & 75.44 & 60.00 & 87.00 & {73.50} \\
    \midrule
     \multirow{16}{*}{\rotatebox[origin=c]{90}{\scriptsize{\textbf{\makecell{Ours (VLM + LLM)}}}}}
     &CLIP~\cite{radford2021learning} \\
     &\quad \quad \small{+ MMed-Llama3-8B-EnIns} & \xmark & 88.17 & 3.71 & 45.94 & 59.41 & 14.61 & 37.01 & 87.00 & 2.00 & 44.50 \\
     &\quad \quad \small{+ Mistral-7B-Instruct} & \xmark & 8.33 & 96.90 & 52.62 & 19.80 & 88.58 & 54.19 & 11.00 & 94.00 & 55.00 \\
    &BiomedCLIP~\cite{zhang2023biomedclip} \\
    &\quad \quad \small{+ MMed-Llama3-8B-EnIns} & \xmark & 21.75 & 87.38 & 54.56 & 21.78 & 70.78 & 46.28 & 21.00 & 70.00 & 45.50 \\
    &\quad \quad \small{+ Mistral-7B-Instruct} & \xmark & 5.00 & 93.59 & 49.29 & 9.90 & 79.00 & 44.45 & 10.00 & 78.00 & 44.00 \\
     &MONET~\cite{kim2024transparent} \\
     &\quad \quad \small{+ MMed-Llama3-8B-EnIns} & \xmark & 100.0 & 28.67 & 64.33 & 89.11 & 31.96 & 60.54 & 77.00 & 45.00 & {61.00}\\
     &\quad \quad \small{+ Mistral-7B-Instruct} & \xmark & 73.17 & 61.30 & 67.24 & 73.27 & 62.56 & 67.91 & 40.00 & 67.00 & 53.50 \\
    &CBI-VLM~\cite{patricio2024towards} & \\
    &\quad \quad \small{+ MMed-Llama3-8B-EnIns} & \xmark & 6.11 & 96.27 & 51.19 & 6.93 & 97.26 & 52.10 & 77.00 & 11.00 & 44.00 \\
    &\quad \quad \small{+ Mistral-7B-Instruct} & \xmark & 86.67 & 0.59 & 43.63 & 99.01 & 11.87 & 55.44 & 100.0 & 0.00 & 50.00\\
    &ExpLICD~\cite{gao2024aligning} \\
   & \quad \quad \small{+ MMed-Llama3-8B-EnIns} & \xmark & 58.49 & 97.65 & {78.07} & 65.35 & 91.78 & {78.56} & 66.00 & 86.00 & \textbf{76.00}\\
    &\quad \quad \small{+ Mistral-7B-Instruct} & \xmark & 72.06 & 82.82 & 77.44 & 67.33 & 90.87 & \underline{79.10} & 72.00 & 79.00 & \underline{75.50} \\
    &\quad \quad \small{+ GPT-4} & \xmark & 72.78 & 93.96 & 83.37 & 67.33 & 90.87 & \underline{79.10} & 77.00 & 72.00 & 74.50 \\
    &  $ \bigstar $ {\small{\textbf{ (ExpLICD + LLM w/ few-shot)}}} & \xmark & 79.44 & 90.65 & \textbf{85.05} & 67.33 & 92.24 & \textbf{79.78} & 76.00 & 74.00 & 75.00 \\
    
    \bottomrule
  \end{tabular}
   \end{adjustbox}
  \caption{\textbf{Disease classification performance across various datasets}. Reported results in \%. The best results are highlighted in \textbf{bold}, and the second-best results are \underline{underlined}.}
  \label{tab:performance_zero_shot_all}
\vspace{-0.3cm}
\end{table}

Table \ref{tab:performance_zero_shot_all} presents the disease classification performance of models in terms of Balanced Accuracy (BAcc), Sensitivity, and Specificity across different datasets in a zero-shot setting. 

\paragraph{\textbf{Zero-shot classification performance}} Using our two-step methodology (VLM + LLM), CLIP-like models such as CLIP and BiomedCLIP generally exhibit poor zero-shot performance, especially in terms of sensitivity, as they frequently predict the benign class (nevus). 
In contrast, Vision-Language Models (VLMs) fine-tuned on dermoscopic images perform significantly better, with MONET and ExpLICD showing the highest accuracy. Although CBI-VLM was also fine-tuned on dermoscopic data, it underperformed compared to MONET and ExpLICD. This may be due to its reliance on image-text alignment only in the final layers of CLIP, which were trained on generic data, while keeping the remaining weights frozen.
When combining ExpLICD with MMed for PH2 and HAM10000, and ExpLICD with Mistral for Derm7pt, we increase BAcc over traditional CBM: 8\% on PH2, 1.5\% on Derm7pt, and 5\% on HAM10000. These gains can be attributed to ExpLICD's use of a pretrained vision-language model and its incorporation of textual diagnostic criteria as knowledge anchors. These anchors are aligned with predefined visual concept tokens through contrastive loss, resulting in more fine-grained alignment than traditional image-text alignment in VLMs. We also evaluate the combination of ExpLICD + GPT-4 due the strong performance of ExpLICD with MMed and Mistral. However, no substantial improvements were observed compared to using open-source LLMs like MMed and Mistral.

\begin{figure*}[h!]
    \begin{tikzpicture}[font=\scriptsize]
        \node[anchor=west] (mistral) at (0,0) {\includegraphics[width=0.5\textwidth]{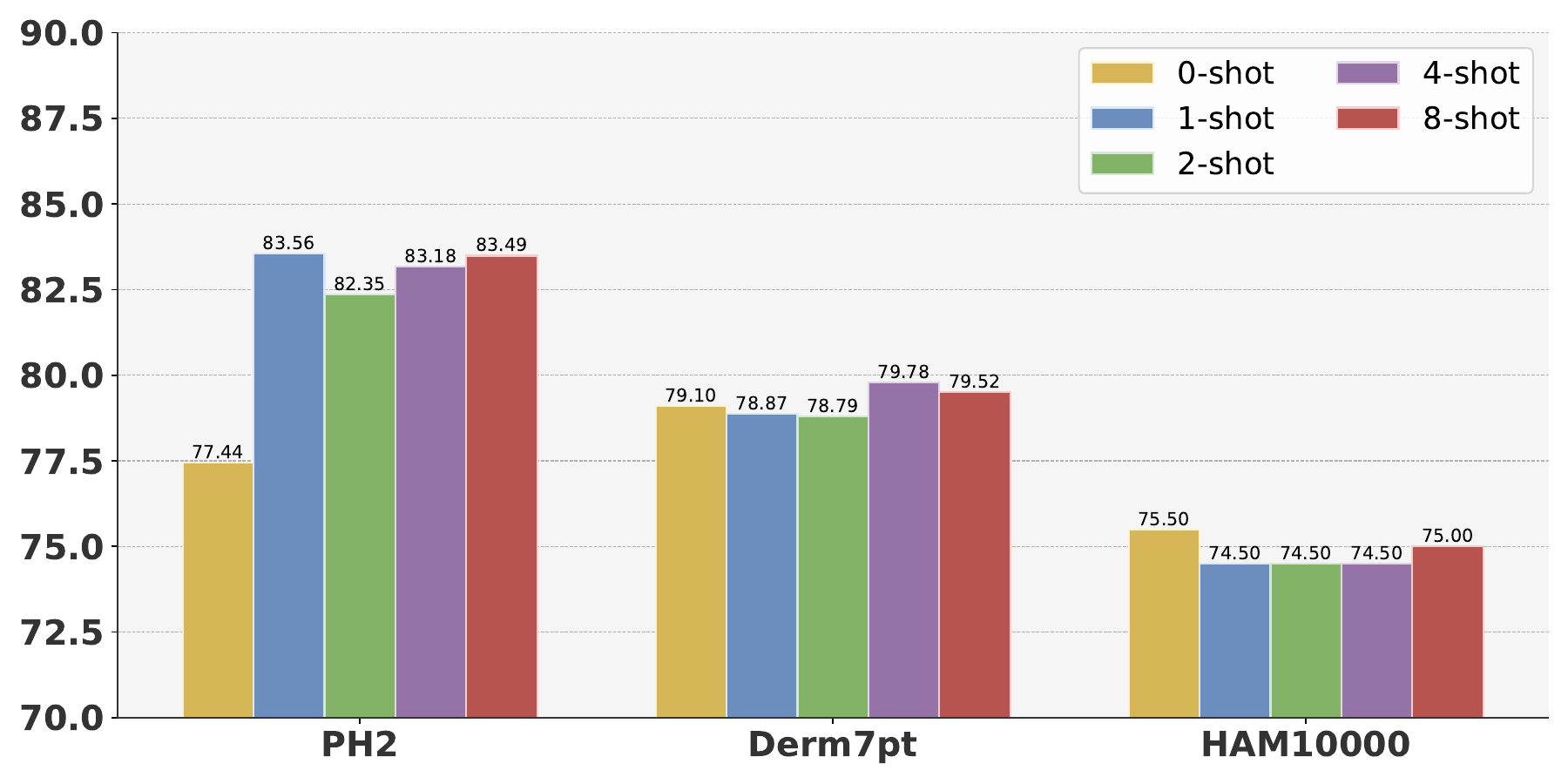}};

        \node[anchor=west] (mmed) at (0.5\textwidth,0) {\includegraphics[width=0.5\textwidth]{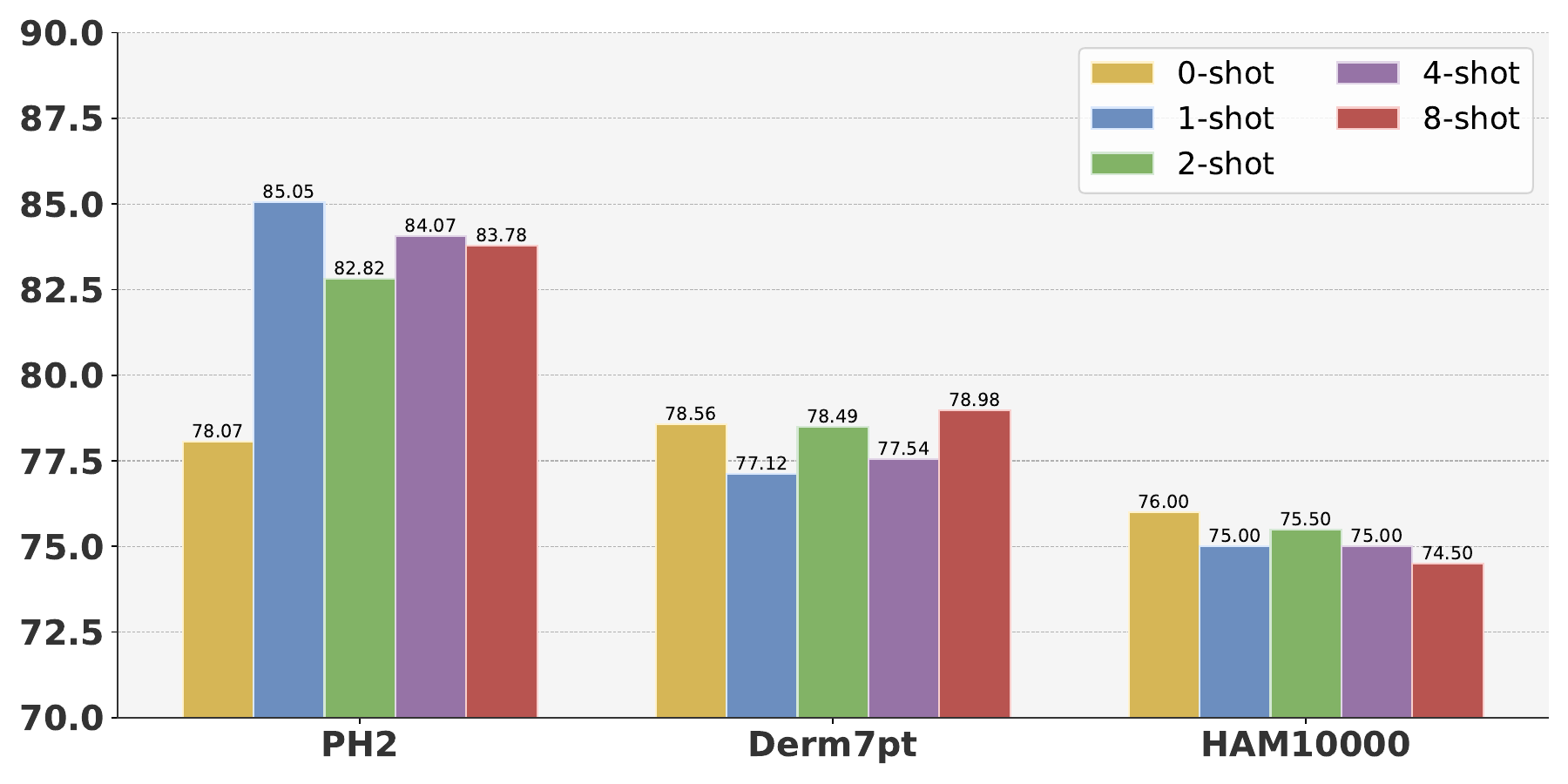}};

        \node at (0.27\textwidth,2) {\textbf{Mistral}};
        \node at (0.77\textwidth,2) {\textbf{MMed}};
    \end{tikzpicture}
    \caption{\textbf{Few-shot disease classification performance (in BACC \%) across different $n$-shot settings}. Each bar corresponds to an $n$-shot scenario $(n = {0, 1, 2, 4, 8})$.}
    \label{fig:plots_few_shot_results}
\end{figure*}

\paragraph{\textbf{Few-shot classification performance}}
Figure \ref{fig:plots_few_shot_results} shows that the results for PH2 and Derm7pt can be further improved by incorporating additional context into the prompt through the inclusion of $n$ demonstration examples, selected using our retrieval strategy. As shown in Figure \ref{fig:plots_few_shot_results}, for PH2, we achieve a score of 83.56\% and 85.05\% when using Mistral and MMed, respectively, with only one demonstration example, significantly outperforming the zero-shot performance. For Derm7pt, the zero-shot performance improves with $K=4$ demonstrations in Mistral and with $K=8$ demonstration examples in MMed. Although no BAcc improvement was observed for HAM10000 using ICL, we achieved increased sensitivity with two demonstration examples in MMed and one in Mistral. 

\subsection{Ablation study on disease classification}

To further validate the effectiveness of our method, we conduct ablation studies on the disease classification strategy. Specifically, we explore direct label prediction from the input image ($x \rightarrow y$) by discarding the intermediate concept prediction step. Instead, we compare image features to short text prompt describing the class, detailed in Table \ref{tab:zero_shot_prompts}.

\begin{table}[!h]
  \centering
  \setlength{\tabcolsep}{5pt}
  \resizebox{0.55\textwidth}{!}{%
  \begin{tabular}{lc}
    \toprule
     \textbf{Model} & \textbf{Template} \\
    \midrule
    BiomedCLIP & \texttt{This is a dermoscopic image of \{class\}}\\
    CLIP & \texttt{This is a dermoscopic image of \{class\}} \\
    MONET & \texttt{This is skin image of \{class\}} \\
    CBI-VLM & \texttt{This is dermatoscopy of \{class\}} \\
    ExpLICD &  \texttt{This is a dermoscopic image of \{class\}} \\
    \bottomrule
  \end{tabular}%
  }
  \caption{\textbf{Prompts used for zero-shot image classification}. The \texttt{\{class\}} is replaced by ``Nevus'' or ``Melanoma'' in our experiments.}
  \label{tab:zero_shot_prompts}
\end{table}

As shown in Table \ref{tab:zero_shot_class_x_to_y}, directly comparing image features with encoded class labels yields lower performance than our proposed method (last row of Table \ref{tab:zero_shot_class_x_to_y}), which first predict the clinical concepts and then uses that information for class label prediction. Nonetheless, CBI-VLM achieves the highest performance across most datasets, benefiting from fine-tuning that aligns image and label features within the same class. These results emphasize the value of an intermediate concept prediction step, enhancing both performance and interpretability.

\begin{table}[!h]
  \centering
  \setlength{\tabcolsep}{3.5pt}
  \begin{adjustbox}{width=0.8\textwidth}
  \begin{tabular}{lccc|ccc|ccc}
    \toprule
     \multirow{3}{*}{\textbf{Model}} & \multicolumn{9}{c}{\textbf{Dataset}} \\
     \cmidrule{2-10}
     & \multicolumn{3}{c}{PH$^2$} & \multicolumn{3}{c}{Derm7pt} & \multicolumn{3}{c}{HAM10000} \\
     & \small{Sens.} & \small{Spec.} & \small{BAcc} & \small{Sens.} & \small{Spec.} & \small{BAcc} & \small{Sens.} & \small{Spec.} & \small{BAcc} \\
    \cline{1-10}
    \noalign{\vspace{0.1ex}}
    \rowcolor{mycolor} 
    \multicolumn{10}{c}{\textbf{\textit{Image to Label} $[x \rightarrow y]$}} \\
    \cline{1-10}
    \noalign{\vspace{0.1ex}}
    CLIP~\cite{radford2021learning} & 100.0 & 0.00 & 50.00 & 93.07 & 7.31 & 50.19 & 91.00 & 4.00 & 47.50 \\
    BiomedCLIP~\cite{zhang2023biomedclip} & 18.89 & 94.11 & 56.50 & 69.31 & 62.56 & 65.93 & 45.00 & 68.00 & 56.50 \\
    MONET~\cite{kim2024transparent} & 78.25 & 88.48 & \underline{83.36} & 55.45 & 88.13 & 71.79 & 86.00 & 44.00 & {65.00} \\
    CBI-VLM~\cite{patricio2024towards} & 40.40 & 74.92 & 77.46 & 82.18 & 73.06 & \underline{77.62} & 32.00 & 99.00 & \underline{65.50} \\
    ExpLICD~\cite{gao2024aligning} & 31.82 & 62.67 & 47.25 & 49.50 & 57.99 & 53.75 & 65.00 & 45.00 & 55.00 \\
    \midrule
    $ \bigstar$ {\small{\textbf{ (ExpLICD + LLM w/ few-shot)}}} & 79.44 & 90.65 & \textbf{85.05} & 67.33 & 92.24 & \textbf{79.78} & 76.00 & 74.00 & \textbf{75.00} \\
    
    \bottomrule
  \end{tabular}
  \end{adjustbox}
  \caption{\textbf{Ablation study on zero-shot classification performance}. The best results are highlighted in \textbf{bold}, and the second-best results are \underline{underlined}.}
  \label{tab:zero_shot_class_x_to_y}
\vspace{-0.3cm}
\end{table}

\begin{table}[!h]
  \centering
  \setlength{\tabcolsep}{3.5pt}
  \begin{adjustbox}{width=0.8\textwidth}
  \begin{tabular}{lcc|c|c}
    \toprule
     \multirow{3}{*}{\textbf{Model}} & \multirow{3}{*}{\textbf{Intervention}} & \multicolumn{3}{c}{\textbf{Dataset}} \\
     \cmidrule{3-5}
     & & \multicolumn{1}{c}{PH$^2$} & \multicolumn{1}{c}{Derm7pt} & \multicolumn{1}{c}{HAM10000} \\
     & & \small{BAcc} & \small{BAcc} & \small{BAcc} \\
    \cline{1-5}
    \noalign{\vspace{0.1ex}}
    \rowcolor{mycolor} 
    \multicolumn{5}{c}{\textbf{\textit{Concepts to Label} $[c \rightarrow y]$}} \\
    \cline{1-5}
    \noalign{\vspace{0.5ex}}

    MMed-Llama3-8B-EnIns~\cite{qiu2024towards} & \xmark & 64.33 & 60.54 & 61.00 \\
    MMed-Llama3-8B-EnIns~\cite{qiu2024towards} & \cmark & \textbf{75.27} {\scriptsize{\improvement{10.94}}} & \textbf{81.09} {\scriptsize{\improvement{20.55}}} & \textbf{80.00} {\scriptsize{\improvement{19.00}}} \\
    \midrule
    Mistral-7B-Instruct & \xmark &  67.24 &  67.91  & 53.50 \\    
    Mistral-7B-Instruct & \cmark & \textbf{73.43} {\scriptsize{\improvement{6.19}}} &  \textbf{72.56} {\scriptsize{\improvement{4.65}}} &  \textbf{58.00} {\scriptsize{\improvement{4.50}}} \\
    
    \bottomrule
  \end{tabular}
  \end{adjustbox}
  \caption{\textbf{Test-time intervention results}. BAcc is Balanced Accuracy. Reported results in \%. The best results are highlighted in \textbf{bold}.}
  \label{tab:testtimeintervention}
\vspace{-0.3cm}
\end{table}

\subsection{Test-Time Intervention}

Correcting mispredicted concepts during inference time is especially valuable in high-stakes fields like medicine. By addressing incorrect concepts, clinicians can help the LLM correcting its final diagnosis. To simulate a scenario where a clinician intervenes to correct erroneous concepts, we replace all predicted concepts with their ground-truth values and evaluate the LLM’s classification performance. We conducted experiments on all datasets using the top-performing CLIP-like VLM (MONET) for comparison.

As presented in Table \ref{tab:testtimeintervention}, intervening on the predicted concepts leads to a significant improvement in Balanced Accuracy across all datasets, with MMed demonstrating the greatest average improvement. These findings underscore the critical role of robust concept detection and emphasize the value of human supervision when applying such methods in clinical practice.

\subsection{Diagnostic Interpretability}

\begin{figure}[htbp]
    \centering
    \includegraphics[width=\textwidth]{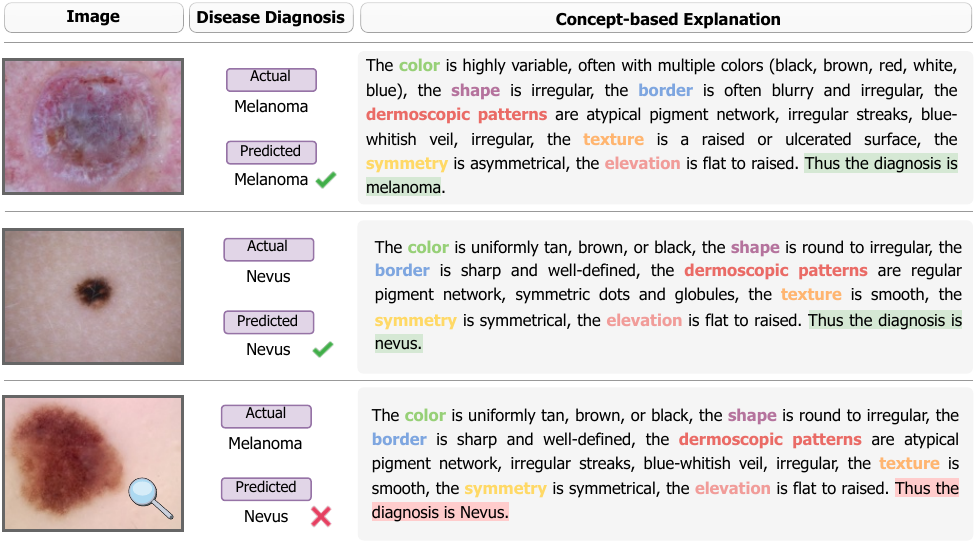}
    \caption{\textbf{Examples of skin images diagnosed using our approach.} The concept-based explanations alongside the predicted disease label allow for further inspection of the skin image, especially when a concept prediction appears inconsistent with the diagnosis.}
    \label{fig:qualitative_results}
\end{figure}

An appealing ability of the proposed methodology is its ability to provide clinical concept-based explanations alongside a diagnosis grounded in those concepts. Figure \ref{fig:qualitative_results} presents two examples of correctly diagnosed skin images along with their concept-based explanations, as well as one example with an incorrect diagnosis. The concept ontology defined in ExpLICD, which align with the ABCDEs of melanoma~\cite{rigel2005abcde}, allow for further examination of the skin image, especially when a predicted concept appears inconsistent with the diagnosis (e.g. the symmetry concept in the bottom row of Figure \ref{fig:qualitative_results}). This two-step approach ensures both interpretability and transparency in the model's decision-making. 

\begin{figure*}[htbp]
    \begin{tikzpicture}[font=\scriptsize]
        \node[anchor=west] (mistral) at (0,0) {\includegraphics[width=0.45\textwidth]{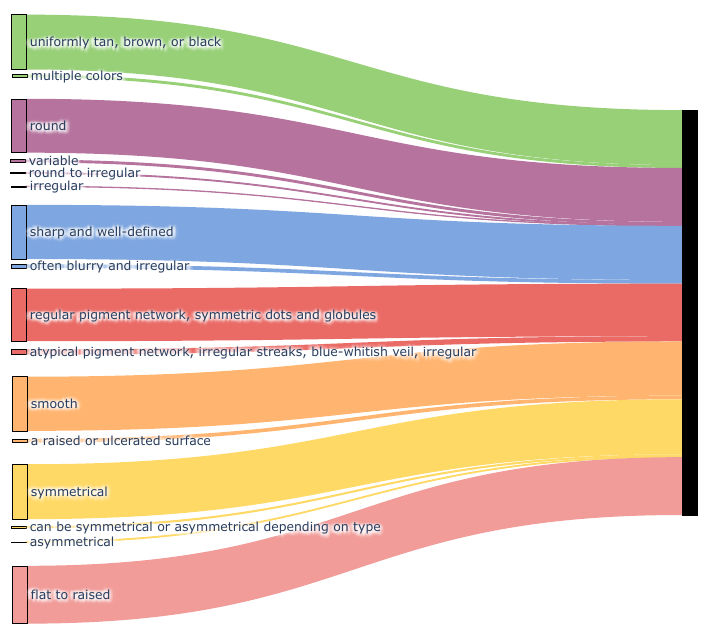}};

        \node[anchor=west] (mmed) at (0.5\textwidth,0) {\includegraphics[width=0.45\textwidth]{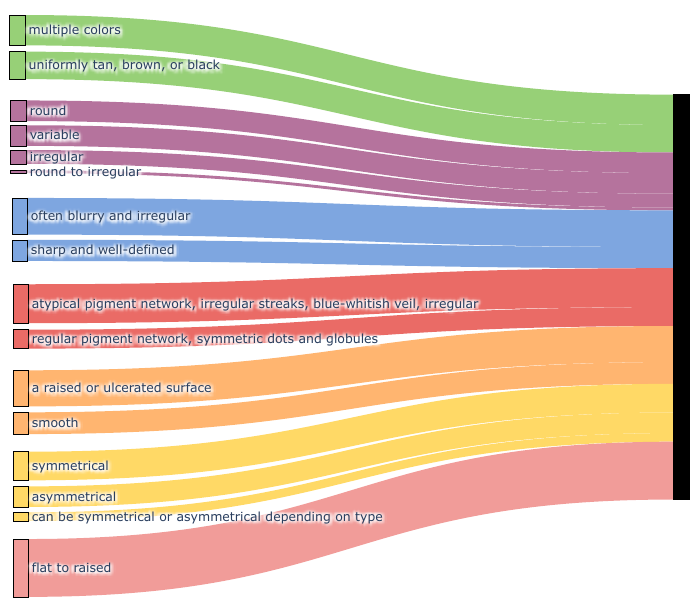}};

        \node[rotate=90] at (0.475\textwidth,0.0) {\normalsize{\textbf{Nevus}}};
        \node[rotate=90] at (0.975\textwidth,0.0) {\normalsize{\textbf{Melanoma}}};

        \filldraw[fill=blue,draw=black] (0.3,-3.55) rectangle (0.6,-3.25);
        \node at (1.05,-3.4) {\scriptsize{{Color}}};

        \filldraw[fill=orange,draw=black] (1.7,-3.55) rectangle (2.0,-3.25);
        \node at (2.5,-3.4) {\scriptsize{{Shape}}};

        \filldraw[fill=green,draw=black] (3.2,-3.55) rectangle (3.5,-3.25);
        \node at (4,-3.4) {\scriptsize{{Border}}};

        \filldraw[fill=red,draw=black] (4.7,-3.55) rectangle (5,-3.25);
        \node at (6.45,-3.4) {\scriptsize{{Demoscopic Patterns}}};

        \filldraw[fill=purple,draw=black] (8.15,-3.55) rectangle (8.45,-3.25);
        \node at (9.05,-3.4) {\scriptsize{{Texture}}};

        \filldraw[fill=brown,draw=black] (9.95,-3.55) rectangle (10.25,-3.25);
        \node at (11.05,-3.4) {\scriptsize{{Symmetry}}};

        \filldraw[fill=pink,draw=black] (12,-3.55) rectangle (12.3,-3.25);
        \node at (13,-3.4) {\scriptsize{{Elevation}}};

    \end{tikzpicture}
    \caption{\textbf{Predicted dermoscopic concepts for each target class in Derm7pt}. The width of the lines indicates the frequency with which each concept is predicted for its respective class. This visualization highlights the key features associated with Nevus and melanoma, providing insights into the most relevant characteristics for each diagnosis.}
    \label{fig:sankey_plots}
\end{figure*}

Figure \ref{fig:sankey_plots} presents the most frequently predicted dermoscopic concepts for each target class in Derm7pt dataset. For Nevus, the predominant predictions include uniform color, well-defined borders, regular pigment network, smooth texture, and symmetry, which are key features typically linked to benign lesions like Nevus. In contrast, for the melanoma class, concepts such as multiple colors, irregular borders, atypical pigment network, ulcerated surface, and asymmetry are the most frequently predicted, and commonly associated with melanoma diagnosis. This analysis offers a transparent and interpretable framework that can be easily understood and validated by medical experts.

\section{{General Discussion}}
\label{sec:general_discussion}

\paragraph{\textbf{The effect of few-shot prompting in the overall performance}}
Our proposed two-step approach consistently outperforms all baseline methods across datasets when using ICL, demonstrating the utility of incorporating additional context into prompts with a few demonstration examples to facilitate LLM reasoning. This improvement was particularly noticeable in the PH2 and Derm7pt datasets. However, no performance gains were observed for HAM10000, which can be attributed to two key factors:
\begin{itemize}
    \item \textbf{Incorrect prediction of concepts}: Our empirical analysis shows that predicted concepts have a significant impact on LLM responses. While demonstration examples can guide the LLM to provide correct diagnoses, situations may arise where incorrectly predicted concepts, such as predicting features commonly associated with melanomas when the ground truth label is nevus, lead the LLM to produce incorrect disease diagnoses, as its response is grounded in the predicted concepts.
    \item \textbf{Inaccuracy in demonstration examples}: Another observed issue is when demonstration examples contain inaccurate associations between concepts and ground truth labels. Since the concepts are automatically predicted by the VLM, there is no guarantee that demonstration examples from the training set accurately represent the correct associations. This mismatch could mislead the LLM into providing incorrect diagnoses.
\end{itemize}
Overall, these issues were frequently observed in HAM10000 dataset, suggesting that it contains more challenging images, which may contribute to a higher False Positive rate. Further refinement of concept prediction and selection of high-quality demonstration examples could mitigate these limitations.

\paragraph{\textbf{Transparency and interpretability in model responses}}
Overall, the results presented in Section \ref{sec:results} highlight the effectiveness of our two-step approach in enabling pretrained VLMs to predict clinical concepts from images and use those concepts to ground the final diagnosis. This enhances both interpretability and transparency in the model's responses, all without requiring additional training and using only a few annotated examples. Notably, our approach outperforms traditional CBMs and state-of-the-art XAI methods across all datasets.

\paragraph{\textbf{Human interaction through test-time intervention}}
Our two-step methodology allows for human intervention in the clinical concepts predicted by the VLM. Since the concepts are presented in plain text to the user, this offers a significant advantage over traditional CBMs, which require adjusting numerical values in the bottleneck layer. Furthermore, our results in Section \ref{subsec:tti} underscore the importance of this test-time intervention in enhancing overall performance.

\paragraph{\textbf{Overall advantages of the proposed two-step approach}}
When comparing our work with studies~\cite{yang2023language,oikarinen2023label,menon2022visual,yan2023robust} using similar techniques, our two-step approach stands out for multiple reasons:
\begin{itemize}
    \item \textbf{Training-free methodology}: We use pretrained models with medical knowledge that require only two inference steps (taking less than 3 seconds) to generate an interpretable final diagnosis. In contrast, existing CBMs and supervised methods require extensive training and are annotation-dependent;
    \item \textbf{Intuitive test-time intervention}: We enable intuitive human intervention to correct any predicted concept by editing plain text;
    \item \textbf{Flexible incorporation of concepts and class labels}: Additional concepts or class labels can be included without the need to retrain the entire architecture. In contrast, CBMs and supervised task-specific models are limited to a predefined set of class labels.
\end{itemize}

\section{Limitations}
\label{sec:limitations}

Although AI-based methods have shown promising results in diagnosing skin lesions, their real-world effectiveness in clinical settings remains largely unexplored. Many studies focus on controlled datasets and scenarios that lack diversity in both diseases and skin tones~\cite{hsiao2024rise}. Therefore, there is a need for the acquisition, annotation, and curation of diverse data, alongside close collaboration between clinicians and AI researchers, to enhance transparency, explainability, and the accuracy of AI integration into clinical practice~\cite{phung2023best}.

\section{Conclusions and Future Work}
\label{sec:conclusion}

We present a novel two-step methodology that utilizes off-the-shelf Vision-Language Models (VLMs) to predict clinical concepts and a Large Language Model (LLM) to generate disease diagnoses based on those concepts. This approach enhances transparency and explainability in decision-making, which is crucial for high-stakes applications like medical use cases. Our evaluation on three skin lesion datasets demonstrates that our method outperforms traditional Concept Bottleneck Models (CBMs), achieving a relative improvement of 21\% on PH2, 3\% on Derm7pt, and 6\% on HAM10000. Additionally, it surpasses state-of-the-art explainable approaches without requiring additional training and uses only a few demonstration examples, thereby addressing the two primary limitations of CBMs. Furthermore, our approach easily incorporates new concepts without retraining and supports test-time interventions. While our methodology was evaluated only on skin image datasets, it is generalizable and can be applied to other domains. In future work, we plan to explore the integration of visual explanations to further enhance trust and promote the adoption of automated diagnosis systems in clinical settings.


\vspace{3mm}
\noindent \textbf{Acknowledgments}\hspace{0.3cm} This work was funded by the Portuguese Foundation for Science and Technology (FCT) under the PhD grant ``2022.11566.BD'', and supported by UID/04516/NOVA Laboratory for Computer Science and Informatics (NOVA LINCS) with the financial support of FCT.IP.

\bibliographystyle{elsarticle-num} 
\bibliography{refs}



\end{document}